\documentclass[sn-mathphys,Numbered]{sn-jnl}

\usepackage{graphicx}%
\usepackage{multirow}%
\usepackage{amsmath,amssymb,amsfonts}%
\usepackage{amsthm}%
\usepackage{mathrsfs}%
\usepackage[title]{appendix}%
\usepackage{xcolor}%
\usepackage{textcomp}%
\usepackage{manyfoot}%
\usepackage{booktabs}%
\usepackage{algorithm}%
\usepackage{algorithmicx}%
\usepackage{algpseudocode}%
\usepackage{listings}%

\usepackage{tabularx}
\usepackage{caption}
\usepackage{longtable}
\usepackage{subcaption}
\usepackage{enumitem}
\usepackage{xurl}
\usepackage{adjustbox}
\usepackage{soul}
\usepackage{tabularray}
\setlist[itemize]{leftmargin=*}

\begin{document}

\title[]{Integrating Large Language Models with Internet of Things Applications}


\author{\fnm{Mingyu} \sur{Zong}}\email{mzong@usc.edu}

\author{\fnm{Arvin} \sur{Hekmati}}\email{hekmati@usc.edu}

\author{\fnm{Michael} \sur{Guastalla}}\email{guastall@usc.edu}

\author{\fnm{Yiyi} \sur{Li}}\email{yiyili@usc.edu}

\author{\fnm{Bhaskar} \sur{Krishnamachari}}\email{bkrishna@usc.edu}

\affil{\orgdiv{Viterbi School of Engineering}, \orgname{University of Southern California}, \orgaddress{\city{Los Angeles}, \state{California}, \country{USA}, \postcode{90089} }}

\abstract{This paper identifies and analyzes applications in which Large Language Models (LLMs) can make Internet of Things (IoT) networks more intelligent and responsive through three case studies from critical topics: DDoS attack detection, macroprogramming over IoT systems, and sensor data processing. Our results reveal that the GPT model under few-shot learning achieves 87.6\% detection accuracy, whereas the fine-tuned GPT increases the value to 94.9\%. Given a macroprogramming framework, the GPT model is capable of writing scripts using high-level functions from the framework to handle possible incidents. Moreover, the GPT model shows efficacy in processing a vast amount of sensor data by offering fast and high-quality responses, which comprise expected results and summarized insights. Overall, the model demonstrates its potential to power a natural language interface. We hope that researchers will find these case studies inspiring to develop further.}





\maketitle

\section{Introduction}
    
    The Internet of Things (IoT) system is a transformational technology in the modern age. It integrates a myriad of devices and enables interconnected devices to communicate and cooperate seamlessly. IoT has been deployed across numerous domains, including transportation, healthcare, and resource management. By harnessing the power of IoT, corporations and individuals can achieve an unprecedented level of automation and real-time supervision with improved efficiency. IoT Analytics reveals that the number of connected IoT devices globally has exceeded 16 billion in the last year \cite{IoT2023}. As the need for more robust and complex networks continues to grow, the research community must address challenges related to system security, smooth device coordination, and efficient data handling to optimize the potential benefits of IoT technology.

    The research community has envisioned that integrating Large Language Models (LLMs) in IoT environments would offer a wide array of advantages that enhance the functionality, intelligence, and automation of IoT systems \cite{Wang2024IoTIT}. LLMs, such as OpenAI's Generative Pre-trained Transformer (GPT) series and Google's Bidirectional Encoder Representations from Transformers (BERT) model, are skilled in natural language processing (NLP) and display an outstanding capability in handling standard benchmark tasks. Additional training cycles allow for continuous comprehensive improvement of these models, keeping them suited for evolving requirements. Moreover, the models are adaptable to downstream tasks through fine-tuning. This flexibility ensures that the models can be tailored to specific needs, not only maximizing the utility of the LLMs but also enabling applications across diverse IoT environments.
    
    A critical benefit of adopting LLMs in IoT environments is the optimization of network security. IoT devices are vulnerable to various cyber attacks due to their limited computational resources for robust security measures. LLMs can contribute to threat mitigation by proposing appropriate countermeasures, such as blocking malicious IP addresses or isolating compromised devices. The Hardware Vulnerability to Weakness Mapping (HW-V2W-Map) Framework for IoT devices employed a GPT model to generate mitigation suggestions for detected system vulnerabilities \cite{Lin2023HWV2WMapHV}. Additionally, LLMs can protect the system by analyzing data patterns and identifying potential security threats in real-time. Among all the cyber threats to an IoT system, Distributed Denial of Service (DDoS) attacks have become a major concern. A DDoS attack involves multiple compromised computers, often spread across locations. By sending a vast amount of requests to the target IoT device, an adversary makes it partially or completely inaccessible to legitimate users, resulting in service interruptions. A recent report by StormWall, a cybersecurity service provider, claims a surge in DDoS attacks across industries over the last year, with a minimum growth rate of 28\% in education \cite{DDoSReport}. The research community primarily relied on machine learning (ML) techniques to detect DDoS attacks before the rise of LLMs. Despite promising performance on the task, the time-consuming nature of training data collection and model re-training posed major challenges for ML classifiers to adapt swiftly to new attack vectors. Moreover, deployment on resource-constrained IoT devices has remained a crucial problem for ML solutions \cite{electronics12143103}. In contrast, LLMs require less time and fewer instances of training to learn new attack vectors. Along with remote access through application programming interface (API), LLMs appear to be potential competitors for DDoS detection. In this study, we explore the possibility of using LLMs for DDoS attack detection in IoT systems. We assessed OpenAI's GPT-3.5, GPT-4, and Ada models on the CICIDS 2017 dataset \cite{CicIDS2017Dataset}. As a result, few-shot learning with 10 examples provided in the prompt allows an LLM to achieve 87.6\% accuracy, while fine-tuning with 70 samples further boosts the number to 94.9\%.

    Large language models further augment IoT environments by utilizing a range of tools, including APIs, programming languages, and integration platforms. By leveraging these tools, LLMs can extend their capabilities beyond natural language processing to include advanced data visualization, real-time monitoring, and automated reporting. Specifically, taking advantage of APIs enables models to integrate data from diverse IoT devices, forming a unified view of the system's state and leading to comprehensive and efficient management of IoT systems. In the realm of virtual embodied environments, the LLM-powered agent Voyager continuously queries information from the Mineflayer \cite{Mineflayer} API and relies on the feedback to explore the Minecraft world \cite{Wang2023VoyagerAO}. On the other hand, seamless and efficient operations over a vast amount of IoT devices require considerable human effort or well-coded scripts. Towards system automation, utilizing tailored scripts has drawn more attention from domain experts. Macroprogramming provides an intermediate interface for managing interconnected IoT devices, thus playing a pivotal role in the field. It depends on high-level programming frameworks that simplify the information retrieval and action execution of applications across distributed systems. In previous works, LLMs have demonstrated proficiency in various code generation assignments. For instance, OpenAI's Codex model is able to find and fix bugs in software, with its performance surpassing traditional automatic program repairing methods \cite{prenner2021automaticprogramrepairopenais}. We further explore LLM's limit in code generation by providing a well-coded IoT macroprogramming framework, PyoT \cite{Azzara2014DemonstrationAP}, and evaluating its understanding and utilization of the framework.

    As IoT networks continue to expand across various industries, efficient and reliable processing of IoT data takes on added importance for the benefits of data storage and data mining. LLMs' advanced ability to interpret and process lengthy natural language input allows them to assist in real-time data handling, capturing patterns that traditional methods might have overlooked. Previous studies have acknowledged their capability in parsing through different forms of IoT data, including logs, alerts, and scripts, to provide concise summaries and actionable insights \cite{Yang2023AutomatedGO, Singla2023AnES, Perrina2023AGIRAC, Fayyazi2023AdvancingTA, Zhang2023CupidLC, Mitra2024LOCALINTELGO, Zhang2024WhenLM}. In this study, we verify the instant capability of GPT-4 model and Gemini-1.5-pro model for processing vast amount of data using IoT sensor datasets and various queries.

    Furthermore, traditional IoT interfaces often require specialized training to manipulate, casting a barrier to widespread employment. However, LLMs incorporate more intuitive interfaces and user-friendly APIs. Through integration of PyoT and data processing tasks, LLMs exhibit potential in sustaining a more accessible interface, enabling users to interact with a complex IoT system through plain natural language queries. This trait also promotes personalized experiences, which boosts user satisfaction and fosters user engagement.

    The key contributions of this paper are:
    \begin{enumerate}[leftmargin=2.5\parindent, rightmargin=2.5\parindent]
      \item Demonstrating the potential of LLMs in optimizing network security by detecting and mitigating cyber threats in real-time.
      \item Highlighting the versatility of LLMs in harnessing the power of macroprogramming frameworks.
      \item Showcasing LLMs' superior performance in IoT sensor data analytics.
    \end{enumerate}

    The rest of this paper is organized as follows: section \ref{related work} presents the related works that have been accomplished in the areas of interest. In section \ref{security}, we illustrate the DDoS detection methodologies that have been utilized, including few-shot learning and fine-tuning techniques. In section \ref{programming}, we prove the efficacy of employing LLMs in a macroprogramming environment. Contributions that LLMs can make to IoT analytics are demonstrated in section \ref{analysis}. A discussion on results and findings is provided in the last section.

\section{Related Work}
\label{related work}

This section introduces key challenges faced by IoT technology and contains three subsections corresponding to the case studies we discuss in this paper: Cyber-threat detection, Macroprogramming for IoT, and  Processing IoT data. A comprehensive summary highlighting the diverse applications of Machine Learning models and Large Language Models in the Internet of Things (IoT) environment is presented in Table \ref{tab:summary}. 

    \subsection{IoT Challenges}
    The IoT technology presents a series of key challenges that complicate its effective implementation and deployment. The complexity of IoT systems, comprised by a mix of devices and protocols, often poses as a bottleneck in ensuring seamless operation and maintenance. Moreover, it has led to low interpretability, which is vital for the systems’ processes and outcomes to be understood and trusted by stakeholders \cite{Gubbi2012InternetOT}. In the era of big data, IoT devices generate vast and diverse streams of information. While collected data provide actionable intelligence, data heterogeneity introduces difficulties in standardization, integration, and analysis, as the varying formats, structures, and semantic meanings of data obstruct seamless interoperability. Since IoT systems must handle and analyze large volumes of data with minimal latency to support timely decision-making, the need for real-time data processing is paramount. However, it is yet to be realized due to the computational demands of rapidly processing vast amount of heterogeneous data \cite{Chen2014BigDA}.

\subsection{Cyber Threat Detection} \label{threatdetection}
    Attacks on IoT systems may target any of the three layers in architecture: perception, network, and application. Early experiments have established the effectiveness of ML techniques as IoT security solutions concerning various dimensions, including access control, authentication, and malware detection \cite{Hussain2024CostOptimizedDA, 9289323, narudin2016evaluation}. In light of LLMs' efficiency and superior performance across diverse tasks, leveraging the power of LLMs for IoT threat mitigation has been an emerging research interest.

    IoT data are known for being private and diverse, making it challenging for a defensive ML model to achieve satisfactory performance, due to the insufficiency of training data and unique characteristics of individual environment. Thus, there is a growing need to synthesize representative IoT data concerning context and a system's requirements, which LLMs can handle independently. In recent work, Song et al.\cite{10248596} proved that a fine-tuned LLM could effectively synthesize high-quality fake cyber threat intelligence (CTI) data, addressing the issue of excessive time consumption in manual data collection.
    
    Cyber threat detection frameworks gain enhanced precision with integration of LLMs. Detection accuracy of advanced persistent threat (APT) with traditional ML networks such as Perceptron and Convolutional Neural Network (CNN) declines as the attack sequence length increases. In contrast, a BERT model's performance is mainly unaffected due to its superiority in processing long sequential data \cite{9627827}. Zhang et al.\cite{Zhang2023IntelligentNT} fine-tuned GPT-2 for high-accuracy real-time IoT attack detection and incorporated the model into a highly adaptive engine. Ferrag et al.\cite{10423646} pre-trained the SecurityBERT model to distinguish fourteen types of cyber incidents from legitimate traffic data. Samples were hashed before being encoded to ensure data security. For many prior ML frameworks, intensive computation and communication costs remain a bottleneck to accommodate resource-constrained devices \cite{xiao2018iot}. However, SecurityBERT is suitable for IoT environments with limited resources, given its lightweight. Despite the benefits of adopting LLMs for threat detection, the nascent stage of deploying LLMs for network security must be highlighted. LLMs are hardly interpretable; ongoing researches struggle to uncover the logic behind model responses. On the other hand, FalconLLM-crafted responses for labeled threats tend to lack the specificity required for individual systems \cite{ferrag2023llm_cyber}. In contrast, our approach aims to harness a pre-trained LLM, not only for attack detection but also to elucidate the reasoning behind threat identification.

\subsection{Macroprogramming for IoT}
    IoT macroprogramming abstracts the system, presenting it as one programmable entity \cite{10.1145/3579353}. Instead of micromanaging every end device, it allows developers to operate on high-level specifications and configure the environment for a set of devices or the entire network. The macroprogramming framework automatically decomposes high-level commands and handles the distribution and execution of sub-tasks. This paradigm is designed to simplify the development and management of large-scale IoT networks, consequently facilitating productivity. Specifically, the innovative framework D’Artagnan \cite{Mizzi2018DArtagnanAE} is constructed at the network level for distributed systems like IoT. Foundational building blocks utilize stream processors whose descriptions can be interpreted, transformed, and analyzed. Obtained information undergoes layers of higher-level abstraction with the help of stream operators to form the final representation . Saputra et al. \cite{Saputra2019WARBLEPA} implemented a middleware Warble to provide simplification for application development, assuming a single application interacts with multiple IoT devices. On the other hand, PyoT \cite{Azzara2014DemonstrationAP} enables high-level management access through rich web interfaces, including the IPython Notebook. Along with the web interface, this macroprogramming framework comprises a control center, a database, and several worker nodes. Each node is responsible for one IoT network, aiming to accomplish seamless management over large-scale systems. It also incorprates an interface for users to create a distinct type of programming abstraction tasks, namely the T-Res tasks. We validate LLMs' value in integrated IoT and macroprogramming environment with concrete use cases of PyoT.

\subsection{Processing IoT Sensor Data}
    The vast amount of IoT sensor data offers substantial value across industries. The value of sensor data processing lies in its ability to transform raw data into critical insights, leading to enhanced operational efficiency and informed decision-making. However, the massive amount of available data poses a primary difficulty to the analysis process, challenging traditional data storage methods and hardware capacity. Given the incentives, numerous efforts have been put into finding solutions that adapt to its high volume. Cloud services and encryptions are among the techniques selected to lighten the burden on IoT data storage \cite{Wang2023SecureAD, agriculture13020274}. Concerning sensor data processing, machine learning algorithms draw specific attention due to their extraordinary capability in pattern identification and summarization. With anomaly detection, occasionally formulated as a classification problem, recent studies have achieved convinsing results with various algorithms, such as K-Nearest Neighbors and Support Vector Machine \cite{DeMedeiros2023ASO, 9822723}. Unsupervised machine learning tasks that require clustering solutions, including IoT data labeling, have been addressed by algorithms like K-means \cite{Yang_2022}. Prior works compared multiple models and summarized the best fit for each task. It is promising that ML models could handle the unprecedented volume and variety of IoT sensor data. Nonetheless, as discussed in \ref{threatdetection}, applying ML models in IoT environments faces the challenges of enhancing robustness and reducing the resources required by deployment. LLMs' remote access empowers them to be superior candidates. LLMs can not only be queried to decipher the raw information but also automatically generate concrete code for carrying out the processing steps. Execution of these code provides additional insightful information and visualizations, tailored to the user's specific needs. This capability marks a significant leap in making IoT data more accessible and actionable. We present a detailed analysis of LLM-assisted IoT data processing in later sections to show LLMs' efficacy in this field.

\begin{tiny}
\begin{longtable}{|p{1.4cm}|p{3.2cm}|p{3.6cm}|p{3cm}|}
\hline
Reference & Description & Pros & Cons \\
\hline
\cite{Hussain2024CostOptimizedDA} & A framework incorporated Artificial Neural Network (ANN) and Support Vector Machine (SVM) for access control. &\begin{itemize} \item Enhanced security measure with pattern recognition and binary classification 
                \item Scalable framework 
                \item Prioritized re-training on most frequently accessed data
\end{itemize} &\begin{itemize}
    \item High computational cost
    \item Binary classification applies to restricted scenarios
\end{itemize} \\
\hline
\cite{9289323} & Evaluation of 24 ML classifiers on biometric data-based authentication. SVM outperformed the other models. & \begin{itemize}
    \item Shed light on model selection for similar settings
\end{itemize}  & \begin{itemize}
    \item Task-specific
\end{itemize} \\
\hline
\cite{narudin2016evaluation} & Evaluation of 5 ML classifiers on mobile malware detection. Bayes network and random forest produced the best results. & \begin{itemize}
    \item Shed light on model selection for similar settings
\end{itemize}  & \begin{itemize}
    \item Task-specific
\end{itemize} \\
\hline
\cite{10248596} & Fine-tuned GPT-Neo for generation of high-quality fake cyber threat intelligence (CTI). &\begin{itemize} \item Demonstrated that LLMs can be effectively used for malicious intentions
            \item Proposed possible solutions for fake CTI detection \end{itemize} &\begin{itemize}
                \item Misuse-related risks and ethical concerns
            \end{itemize} \\
\hline
\cite{9627827} & Pre-training BERT for advanced persistent threat (APT) classification.  & \begin{itemize}
    \item Comparative results from constructed Perceptron, CNN, and Long Short-Term Memory (LSTM) models
    \item Proposed optimization technique for pre-processing long attack sequences
\end{itemize} & \begin{itemize}
    \item Potential Overfitting
\end{itemize}\\
\hline
\cite{Zhang2023IntelligentNT} & Network threat detection framework using fine-tuned GPT-2 model and network logs. & \begin{itemize}
    \item Real-time detection with high accuracy and complex data structure (i.e. JSON format)
    \item Integrated visualization functionality
\end{itemize} & \begin{itemize}
    \item Potential latency by assembly of various components
    \item Security risks associated with cloud-based storage
    \item Resource-intensive
\end{itemize} \\
\hline
\cite{10423646} & A lightweight BERT-based model (SecurityBERT) trained with Privacy-Preserving Fixed-Length Encoding (PPFLE) for multi-class cyber threat detection. & \begin{itemize}
    \item Innovative privacy-preserving technique utilizing labelling and hashing methods
    \item Proposed architecture effectively compressed model to 11 million parameters while keeping model's high performance
\end{itemize} & \begin{itemize}
    \item Possible loss of information when PPFLE is applied to sequential data
    \item High dependency on training data may result in generalization problem
\end{itemize} \\
\hline
\cite{ferrag2023llm_cyber} & A BERT model (SecurityLLM) built from scratch and trained for multi-class cyber threat detection. FalconLLM was adopted to analyze detected threats and suggest security-related solutions. & \begin{itemize}
    \item Comparative analysis on SecurityLLM, ML models, and deep learning models' performance
    \item Incorporated an LLM to provide actionable intelligence based on detection results
\end{itemize} & \begin{itemize}
    \item Complexity in building a transformer model
    \item Extensive training
\end{itemize} \\
\hline
\caption{Applications of ML \& LLM algorithms in the environment of IoT.}
\label{tab:summary}
\end{longtable}
\end{tiny}

\section{Case Study I: Cybersecurity} \label{security}

\subsection{Experimental Setup}

\begin{figure}
        \centering \includegraphics[width=0.7\textwidth]{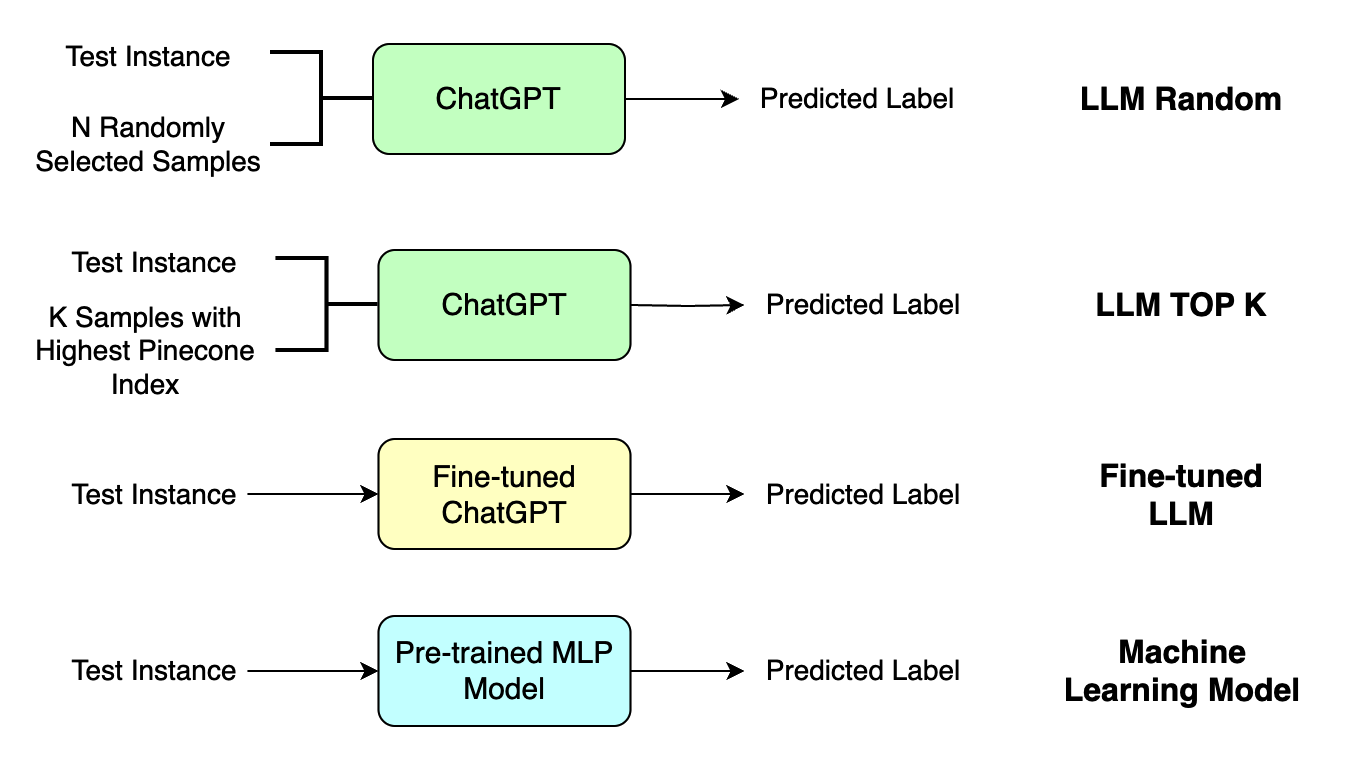}
        \caption{Evaluation Procedures with respect to employed models.}
        \label{fig:flowchart}
    \end{figure}

    The extensive pre-training rendered ChatGPT comprehensive problem-solving skills. We hypothesized that a minimal amount of data is sufficient for the model to understand the context and accurately distinguish DDoS threats from harmless traffic data. To evaluate ChatGPT 4's potential, we adopted both few-shot learning and fine-tuning approaches. Additionally, a traditional machine learning model was trained to verify if LLMs gained reinforced capability over traditional machine learning models in DDoS attack detection (Figure \ref{fig:flowchart}).

    \textbf{Dataset} The CIC-IDS 2017 Dataset \cite{CicIDS2017Dataset} was used for training and evaluation. Every data entry in this dataset has 85 features, along with a ``Benign" or ``DDOS" label. To fit within the limit of model input length, 4 features that convey maximal semantic information have been selected based on results from the previous study \cite{CicIDSFeatureSelection}. In particular, instances were obtained from the ``Friday-WorkingHours-Afternoon-DDOS" pcap file. The processed dataset offers up to 70 training samples.

    \textbf{Few-shot Learning LLMs:} In a few-shot learning setting, LLMs acquire additional contextual information from examples attached to a prompt. This method temporarily promotes a model's ability to understand and handle tasks. This study uses gpt-3.5-turbo to develop two few-shot learning models with subtle distinctions in training data selection.
    \begin{itemize}[leftmargin=0.4in, rightmargin=0.2in]
        \item \textbf{LLM Random} 
        Within each prompt, n randomly selected samples were presented to this model, with n ranging from 0 to 70, and requested it to classify an unseen instance as either "Benign" or "DDOS".
        \item \textbf{LLM Top K}
        Given an unlabelled data, we ranked training samples based on the Pinecone index and retrieved the top k on the list as few-shot examples for this model.
    \end{itemize}

    \textbf{Fine-tuned LLM:} The base model for fine-tuning is the ada model. We formed each training data as a pair of prompt and response. The prompt contained the 4 features of one data point, and the corresponding label was isolated as the expected response.

    \begin{figure}
        \centering \includegraphics[width=0.4\textwidth]{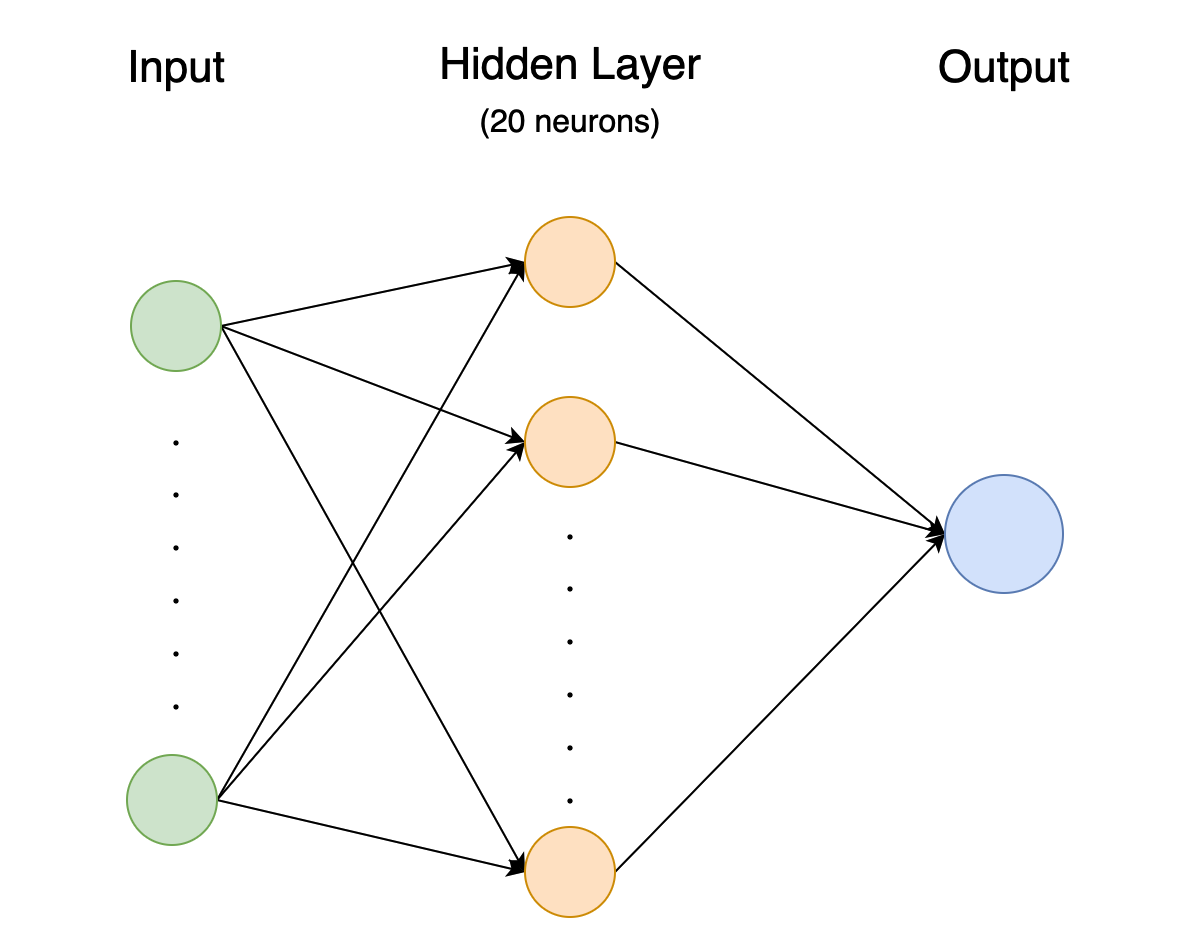}
        \caption{Structure of the MLP model.}
        \label{fig:MLP}
    \end{figure}

    \textbf{Machine Learning Model:} A Multi-layer Perceptron (MLP) \cite{Pal1992MLP} model composed of 1 layer and 20 neurons was trained for comparison (Figure \ref{fig:MLP}). This model was constructed using the ReLU activation function. During evaluation, the model was provided with the same samples that used for few-shot learning models to ensures fair comparison. Since 2 selection methods were involved, we tracked model performance independently for both settings.

    \textbf{Evaluation Metrics:} We used the standard accuracy metric to evaluate models' performance on DDoS attack detection. To gain valuable insights into their behavior, we requested the models to reason about the label of instances. The results were manually reviewed, as no standard measures have been developed for this task.

\subsection{Results}
    Prompt engineering plays a vital role in eliciting high-quality LLM responses. We summarize the findings regarding this process as follows: 
        \begin{itemize}
            \item Instead of presenting only the feature values, pairing each value with its feature name (e.g. Destination Port: 80) resulted in improved detection accuracy.
            \item Explicit definition of separators helps the model understand the structure of prompt. Specifically, each feature was separated by a pipe symbol and each row was separated by a newline in our prompt. A separator of three consecutive \# symbols was used to split training examples and the test case. The uses of all separators were explained to ChatGPT at the beginning of the prompt.
            \item A pre-define output format regulates the model's behavior. Without one, the model is prone to generate responses in various formats, posing problems to automatic evaluation processes. For example, when the model is asked to ``surround the predicted label with `\$\$\$' on each side", it frequently followed the instruction to make predictions, as opposed to refusing to make a prediction when this instruction was left out.
            \item Asking the model to reason over only the features induces observations of data. If a label is provided, the model tends to hallucinate post-hoc reasoning for it, frequently deceiving with the data.
        \end{itemize}

    \begin{figure}[H]
        \centering
        \includegraphics[width=0.7\textwidth]{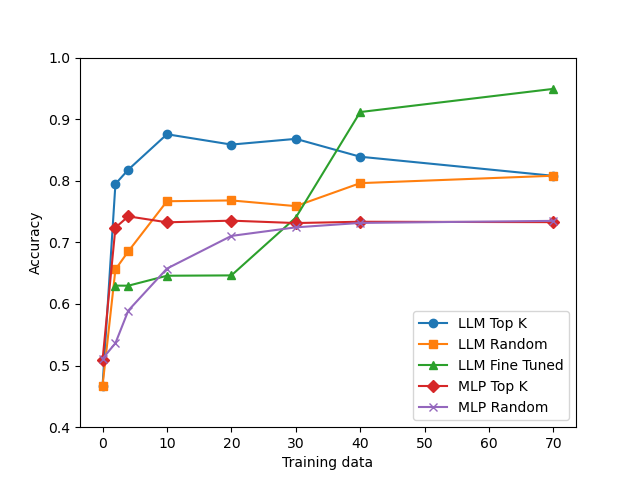}
        \caption{Detection accuracy concerning different models, methods, and training data.}
        \label{fig:few}
    \end{figure}

    Concerning the detection accuracy of DDoS attacks, the best performance was achieved by an LLM solution at most stages. Figure \ref{fig:few} shows fluctuations in highest accuracy with respect to the number of training data. Among the LLMs, our Top K model exhibited advanced potential in addressing this problem with limited samples, whereas the fine-tuned GPT model outperformed the others when more than 40 examples were fed, despite it carrying out the poorest results until the turning point. To be precise, LLM Top K reached 87.6\% accuracy with 10-shot learning, and the model fine-tuned on 70 instances achieved 94.9\% accuracy. Thus, we claim that fine-tuning further boosts the GPT's capability in DDoS attack detection, even though it requires more training data to surpass the performance of a few-shot learning model.

    Following the findings of \cite{liu2023lostmiddlelanguagemodels}, we assumed that the growing size of k might have undermined model performance. These models struggle to keep coherent interpretation as they process through lengthy contents, so they might lose track of focuses. In order to verify this assumption, we made different arrangements for the k samples: a) the most relevant samples were put in the middle, and b) the most relevant samples were located at the beginning and end of the list. However, the resulting detection accuracies are 0.92 and 0.91, respectively. It is worth highlighting that, although some examples are less similar to the test case, they were still among the k most related instances. Therefore, they don't conform to the proposed definition of a ``distractor", consequently caused negligible differences.

    \begin{figure}[H]
        \centering
        \includegraphics[width=0.8\textwidth]{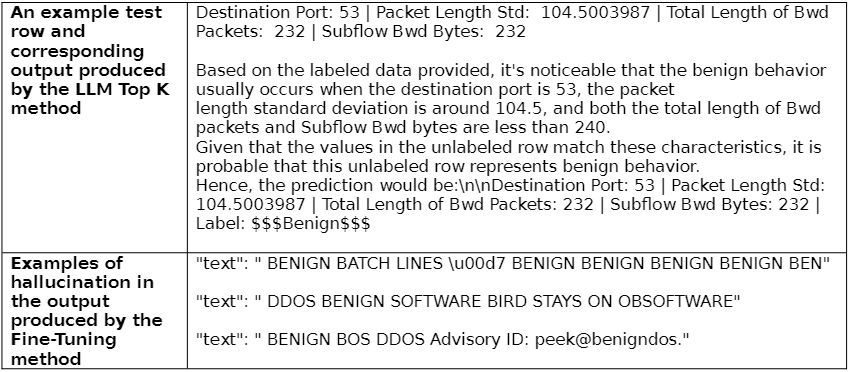}
        \caption{Model explanations of selected labels.}
        \label{fig:llmouts}
    \end{figure}

    We ordered the LLM models to explain why a given instance has the specific label. A GPT-4 model with 20-shot learning was introduced to gather information about latest advancements in GPT models. Without additional training on the reasoning task, the fine-tuned model was more prone to hallucination (Figure \ref{fig:llmouts}). In contrast, both GPT-3.5 and GPT-4 were able to offer valuable explanations for their answers. They were also honest about their uncertainty when they encountered a challenging problem. Specifically, GPT-4 could point out one similar training example of each label. However, it would not consider more related examples and spontaneously utilize the frequency to make a final decision.

\section{Case Study II: IoT Macroprogramming} \label{programming}
Large language models are uniquely positioned to improve the experience with PyoT, a specialized macrogramming language \cite{Azzara2014DemonstrationAP}. Developers can benefit from LLM-enabled automatic code generation and error detection, significantly simplifying the coding process. Furthermore, LLMs can assist in explaining complex PyoT constructs and forming system reports, improving maintainability over complex networks. Integrating PyoT and LLM promises to foster efficient management within the IoT community.

\subsection{Experimental Setup}
    It is believed that a large language model could take advantage of PyoT functionalities to build pipelines for advanced use cases and confirmed this with the ChatGPT-4 model via web interface. Three scenarios were selected for this test; they belong to the most common applications of IoT in the real world: smart home, healthcare, and manufacturing. Description of each scenario and PyoT files were fed into GPT as a single prompt via web interface. The model was asked to output separate scripts for each task and build a comprehensive program to handle all possible incidents. We manually read through the responses to develop perspectives of GPT's behavior. For further improvements to this framework, we interviewed the model to see if it could give novel ideas.
    
\subsection{Results}

    \begin{figure}[H]
        \centering
        \includegraphics[width=\textwidth]{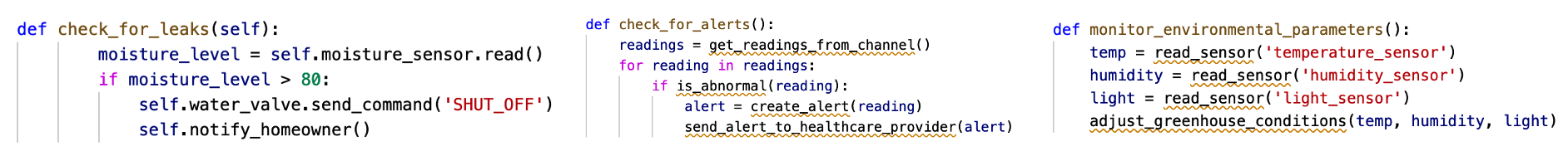}
        \caption{Use cases of retrieved sensor data for smart home (left), healthcare (middle), and agriculture (right).}
        \label{fig:commoncase}
    \end{figure}

    \begin{figure}[H]
        \centering
        \includegraphics[width=\textwidth]{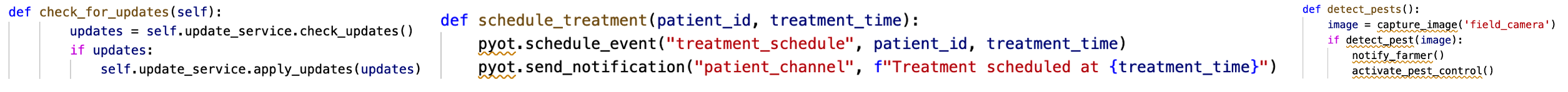}
        \caption{Scenario-driven event-handling for smart home (left), healthcare (middle), and agriculture (right).}
        \label{fig:uniquecase}
    \end{figure}

    \begin{figure}[H]
        \centering
        \includegraphics[width=0.6\textwidth]{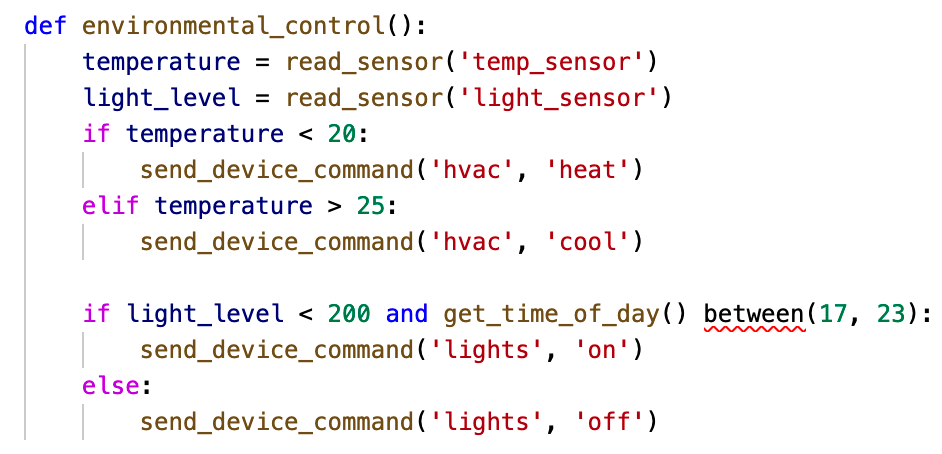}
        \caption{Minor mistake detected in GPT-composed script.}
        \label{fig:typo}
    \end{figure}

    As a result, simple instructions are enough to elicit a list of potential use cases from GPT-4, but only if the model is queried to output the cases separately. Although asking the model to write a complete program saves time in combining code snippets to form a well-structured script, it often fails to consider as many events as covered in the first case. Based on our experience, we would suggest as a best practice to first obtain a list of events and then request the model to assemble all functions in the same chat.

    Given raw scripts, GPT-4 not only is able to identify functionalities of the PyoT framework, but also can adapt them to a specific context. For instance, one universal functionality of PyoT is sensor data retrieval. GPT-4 constructed functions on top of it for water leak detection, equipment alert detection, and environmental monitoring with respect to the given scenarios (Figure \ref{fig:commoncase}). Meanwhile, it suggested unique processes for each of them, tailored to their particular needs. The model implemented automatic checking and installation of updates, which is widely used for modern applications. Under the assumption of the healthcare environment, it would help the patients if a system is able to keep track of scheduled treatment and send out notifications. Real-time pest detection is achievable through IoT systems and image processing. Thus, the model developed a function for this process. The code written by GPT-4 for each case is presented in Figure \ref{fig:uniquecase}.
    
    Besides the tendency to include fewer use cases, the model is prone to make minor mistakes when writing complete programs (Figure \ref{fig:typo}). To alleviate this problem, another round of self-validation or OpenAI's designated model CriticGPT \cite{CriticGPT} could be of assistance. We assume the model has a higher chance of making mistakes as the length of code grows, due to its struggle with coherence, but confirmation of this theory is left for future work.  Moreover, we must emphasize that developed functions could require manual adjustments to be functional within a frame.

    GPT-4 has mentioned several possible contributions it can make to the IoT framework. Some align with previous studies' results, including generating documentation and code snippets; some require further examination, such as finding security issues in the original code and serving as the engine of a natural language interface. A few of them are not even testable, given the current input limit and model configuration. Specifically, if the model is able to dive into a complete project and read through all the documents, we will be able to test its limit on complex tasks such as test case generation and module integration.

\section{Case Study III: Sensor Data Analysis} \label{analysis}

Large language models' web-based interfaces offer considerable benefits for data processing. Integration of supportive components rendered LLMs extended capabilities, such as code execution and visualization, providing accessible platforms for performing complex analytical procedures with minimal domain expertise. However, operations executed to derive results are not completely visible to users, raising questions regarding correctness and reliablity of models' responses. On the contrary, when the task can be fully handled by code, prompting LLMs to write scripts presents a more rigorous approach in terms of transparency, verifiability, and reproducibility. Moreover, users can query an LLM on the web interface, attaching script-generated results, to harness its NLP power and obtain valuable, explainable insights.

\subsection{Experimental Setup}
    This case study tested two large language models' ability in code generation for IoT sensor data processing and analysis. Anand introduced a dataset \cite{sensorTemperature} of temperature readings in a building, which were recorded by IoT devices. An admin could utilize these readings to monitor building temperature and make adjustments. However, processing all the information in real time requires more than human effort. We have crafted 25 instructions for test cases on the temperature dataset, which involve data transformation, data analysis, and construction of visualizations. They were further categorized based on level of difficulty: 10 basic cases, 10 intermediate cases, and 5 advanced cases. Test cases that were considered basic needed simple operations on the input data to derive results. Intermediate level tasks could involve plotting and the use of common knowledge (e.g. concept of the four seasons). So the models were expected to utilize more complicated Python functions. The most advanced tests mostly pertain model fitting (statistical or machine learning), and sometimes require domain knowledge (e.g. heating, ventilation, and air conditioning (HVAC) standards) to handle. We queried the ChatGPT-4 model and the Gemini-1.5-pro model on web interfaces to write Python code for crafted tasks. Each independent prompt contained one test case, a clear instruction to generate complete Python script, and the source file. The occupancy dataset was initially sourced to train classifiers for room occupancy detection \cite{CANDANEDO201628}. It contains more features for each instance, which could potentially serve as distractors when being processed for specific commands. Following the same standards, we formulated 20 distinct queries concerning its content: 5 at the basic level, 10 at the intermediate level, and 5 at the advanced level. And the prompts stayed the same for consistency, except for updates of task description.

\subsection{Results}
    
    A generated script was considered valid if it a) ran successfully and b) addressed all the points in the query. However, both models frequently encountered issues with importing the input files. Among the 25 generated scripts for the temperature dataset, ChatGPT failed to set a correct path and/or name for input file for 23 times. With respect to the occupancy dataset, the occurrence of such mistake was 19. The Gemini model also struggled with file path and/or name, and sometimes hardcoded partial content as a string as if it was all the input data. In total, 39 of the scripts were problematic under the impact. These issues can be resolved through providing additional instructions in prompt, running code on generated results, and manual update. Given the frequency, we decided to eliminate this import factor in order to proceed with the examination on models' potential in terms of code-assisted IoT data processing.

    \begin{table}[h!]
    \centering
    \def\arraystretch{1.2}%
    \begin{adjustbox}{width=\textwidth,center}
    \begin{tabular}{cccc}
    \toprule
    \multirow{2}{*}{\textbf{Query Complexity}} & \multirow{2}{*}{\textbf{ChatGPT-4}} & \multirow{2}{*}{\textbf{Gemini-1.5-pro}} & \textbf{ChatGPT-4} \\
    & & & (with proper handling of feature names) \\ \midrule
    \textbf{Basic} & 5/10 & 7/10 & 8/10 \\ \midrule
    \textbf{Intermediate} & 5/10 & 5/10 & 7/10  \\ \midrule
    \textbf{Advanced} & 1/5 & 2/5 & 2/5  \\
    \bottomrule
    \end{tabular}
    \end{adjustbox}
    \caption{Models' success rates on temperature data analysis.}
    \label{tab:temperatue_tab}
    \end{table}
    
    \begin{table}[h!]
    \centering
    \def\arraystretch{1.2}%
    \begin{adjustbox}{width=\textwidth,center}
    \begin{tabular}{cccc}
    \toprule
    \multirow{2}{*}{\textbf{Query Complexity}} & \multirow{2}{*}{\textbf{ChatGPT-4}} & \multirow{2}{*}{\textbf{Gemini-1.5-pro}} & \textbf{ChatGPT-4}  \\
    & & & (with proper handling of feature names) \\ \midrule
    \textbf{Basic} & 2/5 & 3/5 & 4/5 \\ \midrule
    \textbf{Intermediate} & 3/10 & 9/10 & 6/10  \\ \midrule
    \textbf{Advanced} & 0/5 & 3/5 & 3/5 \\
    \bottomrule
    \end{tabular}
    \end{adjustbox}
    \caption{Models' success rates on room occupancy data analysis.}
    \label{tab:occupancy_tab}
    \end{table}
    
    For temperature-related analysis, ChatGPT-4 and Gemini-1.5-pro produced a comparable amount of valid scripts overall and with respect to each complexity class (Table \ref{tab:temperatue_tab}), whereas the Gemini model's performance on the occupancy dataset surpassed GPT's, especially with harder tasks (Table \ref{tab:occupancy_tab}).
    
    Besides running the scripts, all code and generated results were manually reviewed to determine if the models fully understood their assigned tasks. The process unveiled the cause of GPT's unexpectedly low success rates. Despite the column names being provided in the source file, the GPT model showed a tendency to make up labels for features based on the context. Specifically, one column that stored temperature readings was named ``Temperature [Celsius]", but GPT insisted on referring to it as ``temperature". One categorical variable was ``out/in", indicating if the record was an indoor or outdoor measurement, and ChatGPT used ``Indoor/Outdoor" instead. As a result, the issue was detected in 21 scripts, combining all the test cases. Once they were fixed, 14 more Python scripts would be able to properly handle analysis tasks. On the other hand, ChatGPT displayed higher proficiency in the Python programming language. There were a few occasions where the Gemini model presented misunderstanding of functions and data types, while ChatGPT made no such mistakes throughout the experiment. For instance, when the models were asked to ``calculate a moving average of the temperature with a window of 24 hours", both of them utilized the DataFrame.rolling() function. However, the Gemini model set the rolling window to be 48 rather than 24 because it believed the parameter represented the number of points to include for each window. Another common mistake originated from insufficient understanding of Python would be operations on mixed data types (e.g. subtract a number from a string that appears to be numeric). Regarding the most commonly used Python functionalities, such as plotting and model fitting, both LLMs have been trained to master the corresponding functions and processes. An example test covering statistical analysis and visualization is presented in Table \ref{tab:query_results}. In addition, we have discovered shared difficulties among two models' responses, including processing inconsistent datetime formats and achieving full coverage of intended process, details are listed in Table \ref{tab:issue_summary}.

    \begin{table}[h!]
    \centering
    \fontsize{22}{25}\selectfont
    \def\arraystretch{2}%
    \begin{adjustbox}{width=\textwidth,center}
    \begin{tabular}{ccc}
    \toprule
    \textbf{Query} & \multicolumn{2}{c}{\textbf{Analyze the frequency distribution of temperature values to determine if the distribution is normal, skewed, or has any outliers, using statistical tests and visualizations like histograms and box plots.}} \\ \midrule
    \textbf{Results} & \textbf{ChatGPT-4} & \textbf{Gemini-1.5-pro} \\ \midrule
    \textbf{Distribution} & \raisebox{-0.5\totalheight}{\includegraphics[width=320mm]{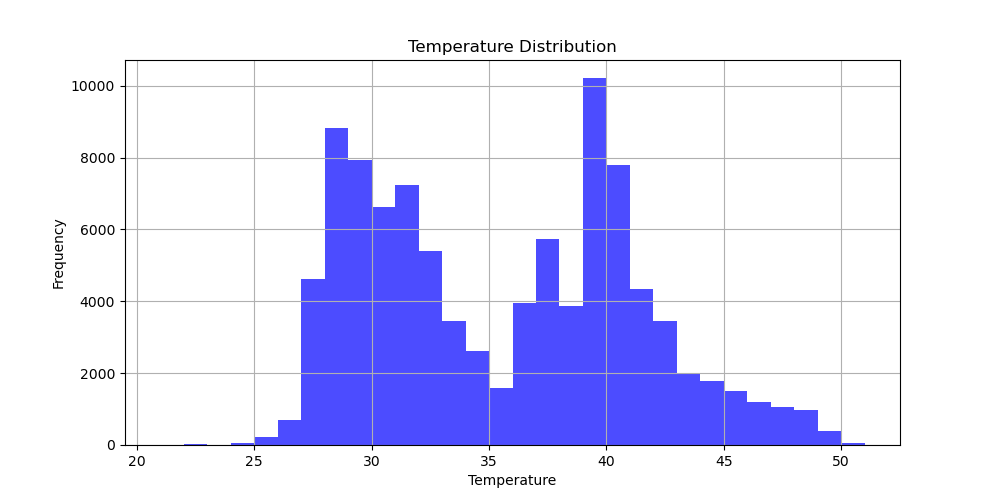}} & \raisebox{-0.5\totalheight}{\includegraphics[height=160mm]{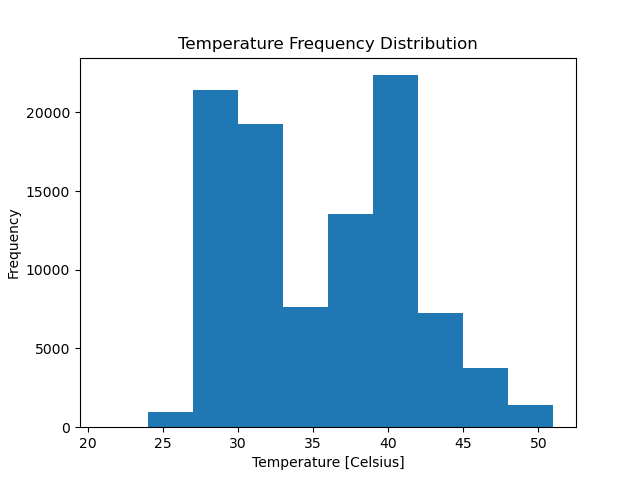}} \\[27ex] \midrule
    \textbf{Outlier Identification} & \raisebox{-0.5\totalheight}{\includegraphics[width=300mm]{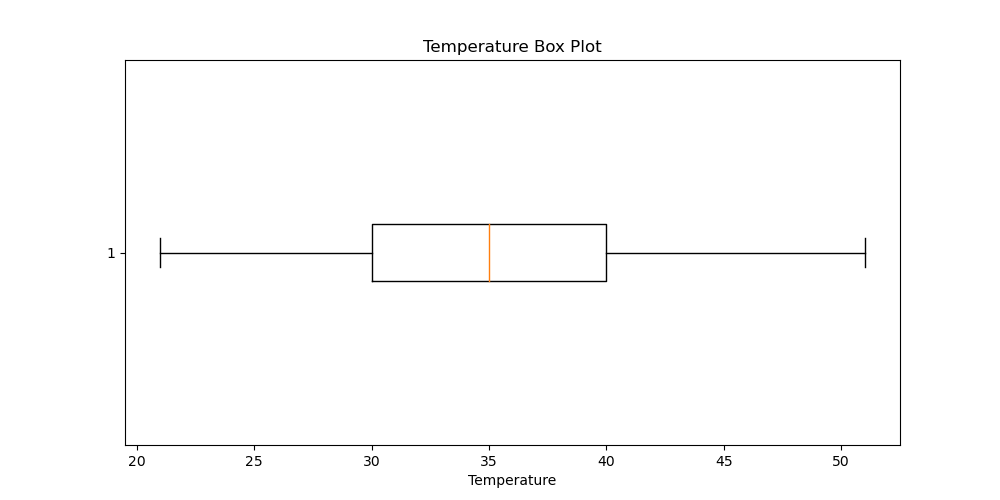}} & \raisebox{-0.5\totalheight}{\includegraphics[height=160mm]{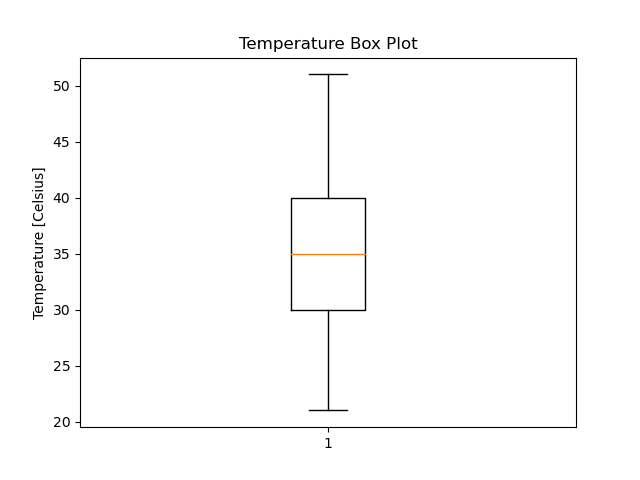}}\\[27ex] \midrule
    \textbf{Textual Information} & \raisebox{-0.5\totalheight}{\includegraphics[width=300mm, height=80mm]{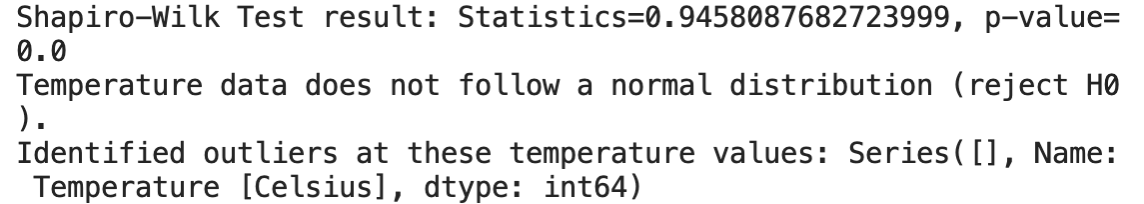}} & \raisebox{-0.5\totalheight}{\includegraphics[height=80mm]{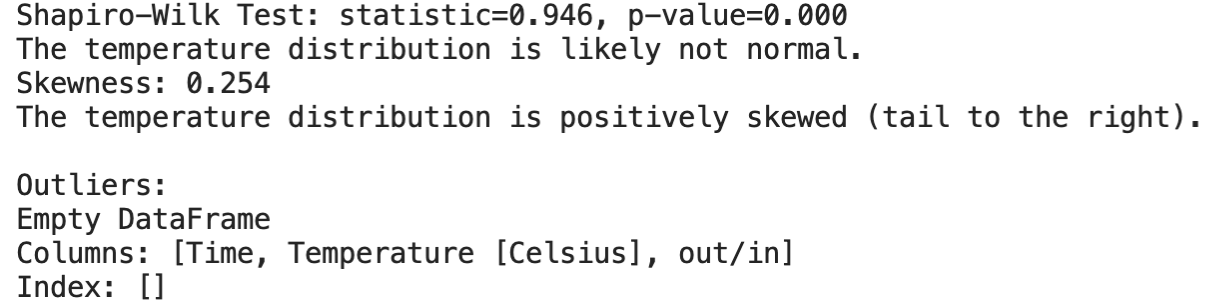}}\\
    \bottomrule
    \end{tabular}
    \end{adjustbox}
    \caption{Example query results.}
    \label{tab:query_results}
    \end{table}

    \begin{table}[h!]
    \centering
    \def\arraystretch{1.2}%
    \begin{adjustbox}{width=\textwidth,center}
    \begin{tabular}{ccccc}
    \toprule
     & \multicolumn{4}{c}{\textbf{Occurrence}} \\ \midrule
     & \multicolumn{2}{c}{\textbf{Temperature Dataset}} & \multicolumn{2}{c}{\textbf{Occupancy Dataset}} \\ \midrule
    \textbf{Problem} & \textbf{ChatGPT-4} & \textbf{Gemini-1.5-pro} & \textbf{ChatGPT-4} & \textbf{Gemini-1.5-pro} \\ \midrule
    Failed data import & 23 & 20 & 19 & 19 \\ \midrule
    Incomplete script & 3 & 1 & 3 & 3 \\ \midrule
    Improper handling of varied datetime formats & 2 & 4 & - & - \\ \midrule
    Wrong column name & 10 & - & 11 & 2 \\ \midrule
    Fabricated column value & 3 & - & - & -  \\ \midrule
    Syntax error & - & - & 2 & - \\ \midrule
    Misuse of function or variable & - & 4 & - & - \\
    \bottomrule
    \end{tabular}
    \end{adjustbox}
    \caption{Common problems detected in scripts.}
    \label{tab:issue_summary}
    \end{table}

    Considering the importance of statistical analysis on IoT data, we attempted to instruct large language models to produce R scripts, since the R language is specialized for statistics and widely used for the purpose. The 10 basic queries on the temperatue dataset were re-used for a pilot study. Unfortunately, the trials could not derive valid results. For GPT-generated code, fabricated labels persistently raised errors during execution. Although the Gemini model was better at plugging in correct feature names, it wasn't equipped with the technique to set up working directory, consequently causing failed loading of input file. Even though the latter issue can be resolved by a simple, automatic process (i.e. inserting a few lines of code at the beginning of each script), both models displayed deficiency in R programming according to coded data manipulations. Therefore, neither model appeared to be competitive under this setting. We hypothesize that R programming-related data was underrepresented during training of LLMs, and the specialized nature of this programming language further compounded such deficiency. But we leave it to be tested by future systematic evaluation.

\section{Conclusion} \label{conclusion}

    In this exploration into the realm of potential applications of large language models for IoT systems, our study has established the efficacy of LLMs in three major areas of application: DDoS threat detection, IoT macroprogramming, and sensor data processing.

    With detailed examination encompassing zero-shot, few-shot, and fine-tuning LLM approaches and a comparative analysis of a conventional multi-layer perceptron (MLP) model, the results showed that LLMs can achieve impressive performance in DDoS detection. Specifically, LLMs approached an 87.6\% accuracy under the few-shot leaning setting. Moreover, the accuracy surged to 94.9\% when the model is fine-tuned on 70 samples. Compared to the MLP model, LLMs consistently showcased superior performance with respect to the growing size of training data. Another valuable insight discovered was the capability of LLMs to articulate the reasons for their decision on DDoS detections. However, fine-tuned LLM has a notable tendency for hallucination, suggesting careful deployment and additional preventative measures to be applied.

    The representative GPT-4 model has demonstrated its ability in utilizing macroprogramming frameworks. The model's instant response and deep understanding of given scripts shed light on a fully automatic management framework of IoT systems through natural language queries. After being trained to process complex projects, LLMs will potentially play a vital role in building such a framework. Nonetheless, our findings suggest that reducing errors in generated code is a critical issue that needs to be addressed for developers to fully trust LLMs, which remains open for future research. Additionally, our experiments only provided specifications of a macroprogramming framework; the IoT system to be integrated is left to the model's imagination. Large language models may need enhanced capabilities to assemble a real IoT network and PyoT to construct valid functions, given added complexity from an IoT network.

    Both test datasets employed for sensor data processing have a considerable size, with more than 97000 and 20000 data points, respectively. Analysis of generated scripts revealed that each model is prone to make a specific set of mistakes, which greatly hindered model's ability to contribute to automatic data processing and analysis. However, those common issues can be identified within a few iterations and fixed through prompt engineering and other simple measures. The finding highlighted the necessity of a systematic evaluation of the chosen LLM before running actual experiments, as it helps unleash the model's full power. Despite the inadequate handling of input file, the selected large language models demonstrated sufficient understanding in Python language and adeptness in dealing with vast, complex datasets, while reserving room for improvement through fine-tuning. Future studies may expand the application of LLMs for real-time data streams from various IoT sensors. Furthermore, a fundamental data processing interface could be developed around an LLM model, harnessing its power in multiple workflow stages.

    As a first work on the topic, we have provided a somewhat high-level exploration of the case studies in this paper. Future work could examine LLMs' application in IoT systems in more detail with quantitative comparisons with other methods or across different LLMs, potentially incorporating different prompt engineering, fine-tuning, and retrieval augmented generation (RAG) techniques to evaluate performance. It merits further investigation of establishing benchmarks for these and related case studies. Explorations of uncovered topics in IoT are also likely to be undertaken by research in the future.






\section*{Declarations}


\begin{itemize}
\item \textbf{Funding}

    This material is based upon work partially supported by Defense Advanced Research Projects Agency (DARPA) under Contract Number\\ HR001120C0160 for the Open, Programmable, Secure 5G (OPS-5G) program. Any views, opinions, and/or findings expressed are those of the author(s) and should not be interpreted as representing the official views or policies of the Department of Defense or the U.S. Government.
    \\
    
\item \textbf{Conflict of interest/Competing interests}

    Not applicable.
    \\

\item \textbf{Ethics approval}

    This work does not involve any human subjects, and no IRB approval was needed. We have used ChatGPT for help with proofreading and editing of the text. The authors accept full responsibility for the contents of the paper. As the paper is about applications of LLMs to IoT, LLMs were also used in the technical evaluations as described.  \\

\item \textbf{Consent to participate}

    Not applicable.
    \\

\item \textbf{Consent for publication}

    Not applicable.
    \\
    
\item \textbf{Availability of data and materials}

    CIC-IDS 2017 Dataset - \url{https://www.unb.ca/cic/datasets/ids-2017.html}

    Sensor Temperature Dataset - \url{https://www.kaggle.com/datasets/atulanandjha/temperature-readings-iot-devices}

    Sensor Occupancy Dataset -  \url{https://www.kaggle.com/datasets/kukuroo3/room-occupancy-detection-data-iot-sensor}

    PyoT Repository - \url{https://github.com/VAPus/PyoT}
    \\
    
\item \textbf{Code availability}

    GitHub Repository - \url{https://github.com/ANRGUSC/LLM_forIoT}
    \\
    
\item \textbf{Authors' contributions}

    Case Study - Cybersecruity: Arvin Hekmati, Michael Guastalla, Yiyi Li

    Case Study - IoT Macroprogramming: Mingyu Zong, Arvin Hekmati

    Case Study - Sensor Data Analysis: Mingyu Zong, Arvin Hekmati

    Paper Writing: Mingyu Zong, Arvin Hekmati, Bhaskar Krishnamachari

    Project Formulation and Supervision: Bhaskar Krishnamachari

\end{itemize}

\bibliography{main}

\end{document}